\documentclass[11pt,a4paper]{article}
\usepackage[hyperref]{acl2020}
\usepackage{times}
\usepackage{latexsym}
\usepackage{graphicx}
\usepackage{subfigure}
\usepackage{amsmath}
\usepackage{multirow}
\usepackage{amssymb}
\usepackage{array}
\usepackage{booktabs}
\usepackage{algorithm}
\usepackage{algorithmic}
\usepackage{makecell}

\usepackage{microtype}

\aclfinalcopy 

\title{When to Talk: Chatbot Controls the Timing of Talking during Multi-turn Open-domain Dialogue Generation}

\author{Tian Lan \\
  Beijing Institute of Technology \\
  \texttt{lantiangmftby@gmail.com} \\\And
  Xianling Mao \\
  Beijing Institute of Technology \\
  \texttt{maoxl@bit.edu.cn} \\}

\date{}

\begin{document}
\maketitle
\begin{abstract}
Despite the multi-turn open-domain dialogue systems have attracted more and more attention and made great progress, 
the existing dialogue systems are still very boring.
Nearly all the existing dialogue models only provide a response when the user's utterance is accepted.
But during daily conversations, humans always decide whether to continue to utter an utterance based on the context.
Intuitively, a dialogue model that can control the timing of talking autonomously based on the conversation context can chat with human more naturally. 
In this paper, we explore the dialogue system that automatically controls the timing of talking during conversation. 
Specifically, we adopt the decision module for the existing dialogue models.
Furthermore, modeling conversation context effectively is very important for controlling the timing of talking.
So we also adopt the graph neural networks to process the context with the natural graph structure.
Extensive experiments on two benchmarks show that controlling the timing of talking can effectively improve the quality of dialogue generation, 
and the proposed methods significantly improve the accuracy of timing of talking.
In addition, we have publicly released the codes of our proposed model\footnote{Github address is https://github.com/xxx(anonymous).}.
\end{abstract}

\section{Introduction}


Building an agent that can chat with humans naturally is a long-term task in the Natural Language Processing which has attracted the attention of a large number of researchers \cite{chen2017survey, Lowe2015TheUD}. Recently, neural approaches have made great progress in building multi-turn end-to-end dialogue systems and lots of neural network models are proposed \cite{Sutskever2014SequenceTS, Serban2015BuildingED, Zhang2018ContextSensitiveGO, Zhang2019ReCoSaDT,}.

\begin{table}[]
    \begin{center}
        \resizebox{0.5\textwidth}{!}{\begin{tabular}{|c|l|}
            \toprule[2pt]
            \hline
            \multirow{8}{*}{\textbf{Context}} & A: Could I see the manager please?                                               \\ \cline{2-2} 
                                              & A: I have a complaint to make                                                    \\ \cline{2-2} 
                                              & B: Yes, I'm the manager here.                                                    \\ \cline{2-2} 
                                              & B: What can I do for you, Madam?                                                 \\ \cline{2-2} 
                                              & A: Did you have the room checked before we move in?                              \\ \cline{2-2} 
                                              & B: Which room are you in?                                                        \\ \cline{2-2} 
                                              & A: \makecell{1808. The toilet doesn't work properly, \\ the water doesn't run in the shower.} \\ \cline{2-2} 
                                              & B: I'm awfully sorry to hear that.                                               \\ \hline
            \textbf{Chatbot Reply}                 & B: \textless{}empty\textgreater{}                                                \\ \hline
            \textbf{Human Reply}                    & B: I'll turn to it right away.                                                   \\ \hline \bottomrule[2pt]
            \end{tabular}
        }
        \caption{In this case, chatbot and human play the role B. The existing chatbots which are trained by alternate examples cannot give a response without user's input. But the human can decide to utter in this time which is more naturally than existing chatbots.}
        \label{tab:10}
    \end{center}
\end{table}

Although researchers proposed lots of novel neural networks models to make the dialogue systems more engaging \cite{Zhang2018ContextSensitiveGO, Zhang2019ReCoSaDT, Hu2019GSNAG}, the existing multi-turn dialogue systems are still very stupid and boring.
Actually most of the existing multi-turn generative dialogue models are trained by paired samples (query, response). So during the conversation, a response can only be obtained when the user's utterances are already provided which is similar with the traditional QA systems \cite{Voorhees1999TheTQ}. This way is very different from the one that the human does. During the daily conversations, human beings always decide whether to utter based on the conversation context.
For example, as shown in Table \ref{tab:10}, the existing chatbots play the role of \textbf{B} and it utters an utterance in the last turn in the conversation context. Without the user's responses to the last utterance, the chatbots cannot continue to provide an utterance and the conversation will be very boring. But during our daily conversations, the human can easily decide to continue to talk based on the context (we denote this ability as controlling the timing of talking). In this case, the conversations are more engaging and natural than the existing chatbots.
Despite this issue can be partially alleviated by using modified the training datasets, the existing dialogue models still cannot decide the timing of talking.
So constructing a dialogue model which can decide when to talk autonomously during conversations is very important and it has not been studied yet.

Our contributions in this work are two-fold:
\begin{itemize}
    \item In order to build a dialogue system that can chat with humans more naturally, we propose a novel generative dialogue model that control the timing of talking autonomously.
    To the best of our knowledge, we are the first one to explicitly control the timing of talking during multi-turn dialogue generation.
    \item To control the timing of talking accurately, we adopt the graph neural networks such as such as GCN \cite{Kipf2016SemiSupervisedCW} and GAT \cite{Velickovic2017GraphAN} to model the natural graph structure of the conversation context.
    Furthermore, we observe that these models do not use the gated mechanism to update the node feature which is inappropriate for the dialogue modeling \cite{Bahdanau2014NeuralMT}, 
    and we propose a novel double-gated graph recurrent neural network to model the timing of talking and generate an appropriate utterance. 
    So during the conversation, our proposed model can effectively decide whether to keep talking or keep silence, just like a normal person.
\end{itemize}

The extensive experiments on two benchmarks show that the quality of dialogue generation can be improved by controlling the timing of talking and 
our proposed graph neural network significantly outperforms the state-of-the-art baselines on the accuracy of the timing of talking. 

\section{Related Work}
\subsection{Dialogue systems}
Dialogue system aims to build an intelligent agent that can chat with human beings naturally which is a important and long-term task in Natural Language Processing and even artificial intelligence \cite{chen2017survey, Lowe2015TheUD, Serban2015BuildingED}. It's useful in a wide range of applications such as virtual assistant and entertainment \cite{young2013pomdp, shawar2007chatbots}.
Dialogue systems can be simply divided into two categories \cite{chen2017survey}: (1) Open-domain dialogue systems, also known as chatbots, have daily conversation with humans; (2) Task-oriented dialog systems, assist humans to accomplish a specific task through conversation. In this paper, we only focus on the generative open-domain dialogue systems.

Since the sequence to sequence architecture \cite{Sutskever2014SequenceTS} was proposed, the end-to-end generative dialogue systems are becoming increasingly popular. Lots of models are proposed to improve the performance of the seq2seq based neural networks approaches \cite{Bahdanau2014NeuralMT, Li2015ADO}. 
As for multi-turn dialogue generation, \cite{Serban2015BuildingED} proposed HRED which uses the hierarchical encoder-decoder framework to model the context sentences. Since then, the HRED based architecture is widely used in multi-turn dialogue generation and lots of variants have been proposed such as VHRED \cite{Serban2016AHL} and CSRR \cite{Shen2019ModelingSR}.

However, all the existing multi-turn open-domain dialogue systems are trained by the pairs of query and response. In this setting, the response can only be obtained when the user provides an utterance. This way is different from the way that human does. So in order to make the multi-turn dialogue systems become more natural and real, we propose a generative dialogue systems that can control the timing of talking during the conversation.

\subsection{Graph neural networks}
More and more learning tasks require dealing with graph data which contains rich relation information among elements \cite{Zhou2018GraphNN}. Graph neural networks (GNNs) \cite{Zhang2018DeepLO} is a very powerful tool to capture the dependence of graphs via message passing between the nodes of graphs and lots of methods are proposed \cite{Velickovic2017GraphAN, Kipf2016SemiSupervisedCW}.
In NLP, GNNs have also become very popular and lots of work start to apply GNNs to model relation information of the semantic units and achieve better performance such as text classification and relation extraction \cite{Peng2018LargeScaleHT, zhang2018graph}. But how to apply the GNNs to generative multi-turn dialogue systems has not been studied.

In this paper, we leverage the role information to effectively understand the context and improve the accuracy of the timing of talking.

\section{Technical Background}
\subsection{Gated Recurrent Unit}
A GRU \cite{Bahdanau2014NeuralMT} and a long short-term memory (LSTM) \cite{hochreiter1997long} are proposed to solve the gradient vanishing existed in the vanilla RNN and both of them have the gated mechanism.

A GRU can be described as:
\begin{equation} \label{form:1}
    \begin{split}
        z_t &=\sigma (W_z\cdot [h_{t-1}, x_t])\\
        r_t &=\sigma (W_r\cdot [h_{t-1}, x_t])\\
        \widetilde{h_{t}} &= tanh (W\cdot [r_t * h_{t-1}, x_t])\\
        h_t &= (1-z_t)*h_{t-1} + z_t * \widetilde{h_t}
    \end{split}
\end{equation} where the $\sigma$ is the \textit{sigmoid} function, $W, W_r, W_z$ are the parameters of the GRU. $h_{t-1}$ is the last hidden state in the RNN and $x_t$ is the input in current step $t$. $z_t$ controls how much past information affects the current state, called the update gate. Compared with the LSTM , the GRU has less parameters and runs faster, so in this paper, we choose to use the GRU instead of the LSTM.

In our work, the gated mechanism is very essential for our proposed graph neural network. It can be used as a method to control the noise of the context and update the state effectively.

\subsection{Graph Neural Network}
A graph neural networks can be described as \cite{fey2019fast}:
\begin{equation} \label{form:6}
    \begin{split}
        h_i^k=\gamma^k(h_i^{k-1}, \delta_{j\in \mathcal{N}_i}(\phi^k (h_i^{k-1}, h_j^{k-1})))
    \end{split}
\end{equation} where $\phi$ denotes differentiable functions such as MLPs (Multi Layer Perceptrons), $h_i^k$ is the node feature of node $i$ in layer $k$, $\mathcal{N}_i$ contains all the neighborhood nodes of node $i$. $\delta$ is the aggregator which collects the information of the neighborhood such as \textit{sum} and \textit{mean}. $\gamma$ is the updator which updates the node feature based on the aggregator's output and the current state $h_i^{k-1}$.
Most of the exisiting graph neural networks implement the $\gamma$ and $\phi$ with the simple linear projection such as GCN and GAT which cannot update the node feature effectively during the dialogue modeling. So we use the GRU to introduce the gated mechanism to alleviate this issue. We compare GCN and GAT with proposed gated graph nerual network in the experiment section.

\section{Notations and Task Formulation}
Let there be 2 speakers: (1) agent: $A$; (2) user: $U$ in a conversation. It should be noted that the motivation of controlling the timing of talking can be easily extended to multi-party conversations, here we only focus on two speakers. Given a list of the utterances as the conversation context $c=\{u_1, a_2, u_3, a_4, ..., u_i\}$, where utterance $u_i$ or $a_i$ is uttered by $U$ or $A$ in the $i$ turn. The goal of the     traditional multi-turn dialogue models is to generate an utterance $r$ that maximizes the conditional likelihood based on $c$:
\begin{equation} \label{form:1}
    \begin{split}
        r &=\mathop{\arg\max}_{r}\sum_{i=1}^{|r|}{\log P(r_i|c,\theta,r_{<i} )}
    \end{split}
\end{equation} where $\theta$ is the parameters of $A$.

In this paper, we formulate the decision of timing of talking during the dialogue generation as a binary classification task. Specifically, given the context $c$ and the speaker information, the model needs to give a decision $t$ to control whether to speak or keep silence at this time.

Then the goal of the dialogue generation with timing of talking is to decide the timing of talking $t$ and generate the response $r$ that maximizes the conditional likelihood given the context $c$:
\begin{equation} \label{form:2}
    \begin{split}
        (t,r) &=\mathop{\arg\max}_{(t,r)}{\log P(t|c,\theta)+\log P(r|t,c,\theta)}
    \end{split}
\end{equation} where $P(r|t,c, \theta)=\Pi_{i=1}^{|r|} P(r_i|t,c,\theta ,r_{<i} )$ is the probability of generating the response $r$ (when the models decide to keep silence, the response $r$ is a special token $\rm{silence}$). Obviously, the dialogue generation with timing of talking is multi-task learning. In our work, the context $c$ is a directed acyclic graph $\mathbf{G}(V,E)$ that contains the utterances and the relations among them, where $V$ is a set of $m$ vertices $\{1,2,...m\}$ and $E=\{e_{i,j}\}_{i,j=1}^m$ is a set of directed edges. Each vectex $i$ is a sentence representation $h_i$ learned by an RNN. If utterance $j$ is related to utterance $i$, then there is an edge from $i$ to $j$ with $e_{i,j}=1$; otherwise $e_{i,j}=0$. The details of constructing the edges can be found in the next section.

\begin{figure*}[h]
    \center{\includegraphics[width=16cm, height=5cm]{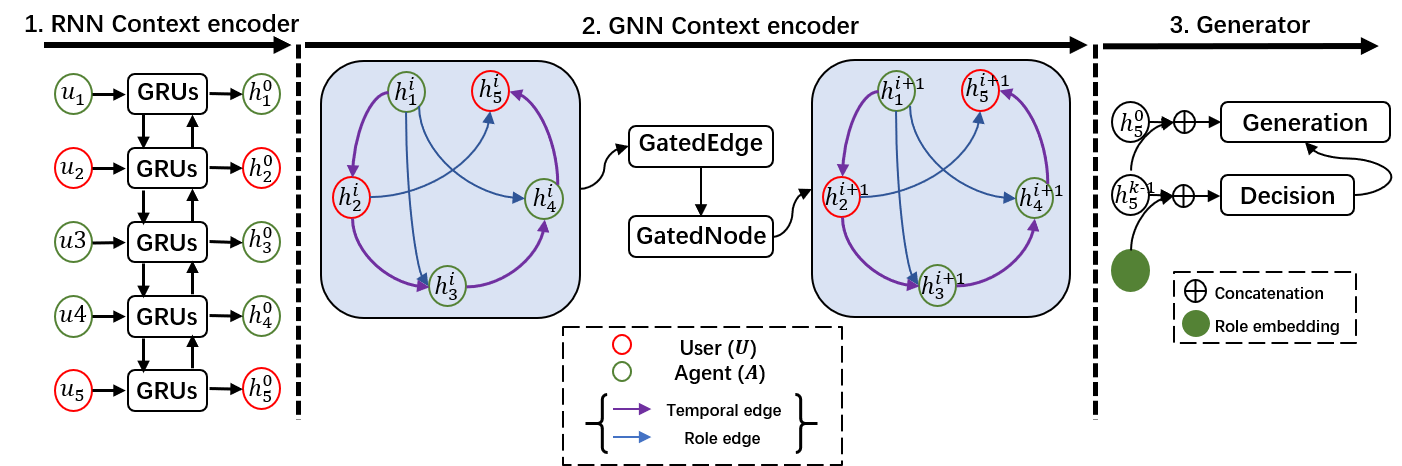}}
    \caption{Overview of the proposed model. In this examples, the conversation has 5 utterances. First, the RNN are used to capture the sentence embedding $u_i$ of the $i$ sentence. Then the Sequential context encoder the Graph context encoder which contains $k$ layers are used to get contextual embeddings $h_i^k$. Finally, in the generator, the model makes the decision of timing of talking and generates an utterance based on $h_i^k$.}
    \label{img:2}
\end{figure*}

\section{Methodology}
In this section, we first demonstrate the way of constrcuting the directed acyclic graph. Then, we show four components of the proposed model: (1) Utterance encoder; (2) Sequential context encoder; (3) Graph context encoder; (4) Generator. 
The overview of the proposed method is illustrated in Figure \ref{img:2}. 

\subsection{Construct the Graph}
As shown in Figure \ref{img:2}, in order to construct an appropriate graph for modeling the timing of talking accurately, we use two kinds of information: (1) Temporal information, it records the topic changes of the conversation as it progresses. We add an directed edge $e_{i-1,i}=1$ between each group of adjacent sentences $(i, i+1)$ to hold this information.; (2) Role information, it's important to make sure your role in the conversation before deciding whether or not to speak. Then we add the edges following: 
\begin{equation} \label{form:3}
    \begin{split}
        e_{j,i}=1, j\in \mathcal{N}_{[0,i-2]}
    \end{split}
\end{equation} where $\mathcal{N}_{[0,i-2]}$ contains all the utterances before $i$ that have the same speaker as $i$.\footnote{We already consider about the relation $e_{i-1,i}$ in the temporal information.}. The details of the graph can be found in experiment section.

\subsection{Double-Gated Graph Recurrent Neural Networks}
\subsubsection{Utterance Encoder}
Given an utterance $i=(w_{i,1}, w_{i, 2}, ..., w_{i, n})$, utterance encoder encodes it into a dense vector $u_i$ called sentence embedding. In this paper, the utterance encoder adopts a bidirectional gated recurrent unit (GRU):
\begin{equation} \label{form:4}
    \begin{split}
        \overrightarrow{u_{i,t}} = \overrightarrow{{\rm GRU}}(\mathbf{e}_{i,t}, \overrightarrow{u_{i,t-1}})\\
        \overleftarrow{u_{i,t}} = \overleftarrow{{\rm GRU}}(\mathbf{e}_{i,t}, \overleftarrow{u_{i,t+1}})
    \end{split}
\end{equation} where $\mathbf{e}_{i,t}$ is the embedding of $w_{i,t}$. The final sentence embedding $u_{i}$ is the concatenation of the last hidden state $[\overrightarrow{u_{i,t}}; \overleftarrow{u_{i,t}}]$.

After encoding by the utterance encoder, the conversation context can be represented with $\mathbf{C} = \{u_i,i\in \{1,2,...,m\}\}$.

\subsubsection{Sequential context Encoder}
Conversations are sequential but all the sentence embedding in $\mathbf{C}=\{u_i,i\in \{1,2,...,m\}\}$ is independent. So, we feed the conversation to an another bidirectional GRU to provide rich contextual information for generator and Graph context encoder:
\begin{equation} \label{form:5}
    \begin{split}
        \overrightarrow{h_{t}^0} = \overrightarrow{{\rm GRU}}(u_{t}, \overrightarrow{u_{t-1}}), t\in \{1,2,...,m\}\\
        \overleftarrow{h_{t}^0} = \overleftarrow{{\rm GRU}}(u_{t}, \overleftarrow{u_{t+1}}), t\in \{1,2,...,m\}
    \end{split}
\end{equation} The contextual sentence embedding of the utterance $i$ is represented with $[\overrightarrow{h_t^0}; \overleftarrow{h_t^0}]$. $\mathbf{C}=\{h_i^0\, i\in \{1,2,...,m\}\}$ is the conversation context.

\subsubsection{Graph context Encoder}
Here, we demonstrate our double-gated graph neural network. First of all, making sure the speaker of each sentence is very important for understanding the utterances. We concatenate the $\mathbf{C}$ and the trainable user embedding to achieve this goal.
Furthermore, during dialogue modeling, the neighbors' positions in the conversation context are beneficial for the aggregator of graph neural networks. But exisiting graph nerual networks ignores it. We also concatenate the $\mathbf{C}$ and the trainable position embedding to address this issue. The final conversation context embedding can be descibed as:
\begin{equation} \label{form:5}
    \begin{split}
        \mathbf{C}=\{[h_i^0; U_{u_i}; P_{u_i}], i \in \{1,2,...,m\}\}
    \end{split}
\end{equation} where the $U_{u_i}$ and $P_{u_i}$ represent the user embedding and the position embedding of the utterance $i$.

As for the sentence $i$ in the conversation context, there may be lots of neighbors when the conversation continues for some rounds because of role information. Not all the neighbors can provide the valuable information for $i$. Some of these neighbors will introduce noises for the node $i$. The existing graph neural networks which treat neighbors equally are inappropriate for the dialogue modeling. Here, we alleviate this problem by applying an GRU to implement $\phi$ of the graph neural network. This is the first gated mechanism of our proposed method. The reset gate of a GRU can control the information flowing from $h_j$ to $h_i$ \cite{Bahdanau2014NeuralMT} to decrease the effect of the noises. The attention mechanism maybe another way to detect the noise and we compare the attention-based graph neural network (GAT) with our proposed model in the experiment section. Finally, the $\delta$ is implemented by \textit{mean} function to aggregate all the information of the neighbors. 

After collecting the neighbors' information by aggregator, the node feature $h_i$ should be updated based on the current $h_i^{k-1}$. The existing graph neural networks use the linear projection to as the updator. But in multi-turn conversations, the gated mechanism can adaptively capture dependencies of different time scales. So we implement update function $\gamma$ with an another GRU. This is the second gated mechansim of the proposed model.

There are $k$ layers in the Graph context encoder which means that each utterance can consider about the neighbors in $k$-steps, and the parameters of these two GRU models are shared across these layers. The final formulation of our proposed double-gated graph neural network can be descirbed as:
\begin{equation} \label{form:6}
    \begin{split}
        h_i^k= {\rm GRU} (h_i^{k-1}, \frac{1}{|\mathcal{N}_i|}\sum_{j\in \mathcal{N}_i}({\rm GRU}(h_i^{k-1}, h_j^{k-1})))
    \end{split}
\end{equation} where the $h_i^{k-1}$ is the node feature of the layer $k-1$. $h_i^k$ is the updated sentence embedding. $\mathcal{N}_i$ contains a set of the neighbors of node $i$.

\subsubsection{Generation and Decision}
As for the multi-turn dialogue generation, the existing dialogue models generate a response. In our work, our proposed model also need to decide whether to talk in this time. If the last utterance is uttered by the user, the model should learn to decide to speak now. If the last utterance is uttered by the model, it needs to decide whether to continue speaking or not according to the context.

As shown in Figure \ref{img:2}, decision module in the generator classifies the timing of talking based on the model's user embedding and the context embedding of the last sentence $h_m^{k-1}$. We implement the decision module by a three layers MLPs with the dropout mechanism \cite{Srivastava2014DropoutAS}, the hidden layers of MLP use $\textit{relu}$ as the activation function, whereas the last unit uses $\textit{sigmoid}$ function:
\begin{equation} \label{form:6}
    \begin{split}
        t = {\rm MLPs}([h_m^{k-1}; U_{model}])
    \end{split}
\end{equation} 
where the $U_{model}$ is the user embedding of the proposed model. It should be noted that the decision model can be easily updated by using reinforcement learning algorithms. Due to page limitations, we leave this part for the future research.

Generation module is a GRU decoder. To generate a response $r$, the decoder calculates a distribution over the vocabulary and sequentially predicts word $r_i$ using a
\textit{softmax} function based on context and decision output $t$:
\begin{equation} \label{form:6}
    \begin{split}
        p(r|h_m^0;h_m^{k-1};\theta) &= \Pi_{j=1}^{|r|}P(r_j|h_m^0;h_m^{k-1};\theta;r_{<j };t)\\
        &= \Pi_{j=1}^{|r|} \textit{softmax}({\rm GRU}(r_{j-1}, \mathbf{h_j})))
    \end{split}
\end{equation} where $r_{j-1}$ is the token generated at the ($j-1$)-th time step, obtained from the word look-up table. $\mathbf{h_j}$ is the hidde state of a GRU at last time step and the initial hidden state is the concatenation of $t$ and context embeddings. $\theta$ is the parameters of the GRU decoder.

\section{Experiments}

\subsection{Datasets}

\begin{table}[t]
    \begin{center}
        \resizebox{0.5\textwidth}{!}{
            \begin{tabular}{|l|c|c|c|c|}
            \toprule[2pt]
            \hline
            \multicolumn{1}{|c|}{\textbf{Dataset}} & \textbf{size} & \textbf{avg turn} & \textbf{degree} & \textbf{avg edges} \\ \hline
            \textbf{Dailydialog}                   &  135/5/5      & 8.82     &  3.91/2.33  & 26.93   \\ \hline
            \textbf{Ubuntu}                        &  59/2/2       & 5.34     &  2.17/1.67  & 9.94     \\ \hline \bottomrule[2pt]
            \end{tabular}
        }
        \caption{The First column contains the train/test/dev data size of each dataset, in thousands. The third column contains the average in-dgree and out-degree in the directed acyclic graph.}
        \label{tab:1}
    \end{center}
\end{table}

It's difficult to evaluate the performance of the proposed models and baselines because most of the existing dialogue datasets only contains the pairs of query and response. The models trained by these datasets ignore the timing of talking and can only generate one response at a time.
We find that the Ubuntu corpus \cite{Lowe2015TheUD} contains the dialogue with the timing information and we use the some conversations of it. In order to make the experimental results more convincing, we also convert the Dailydialog \cite{Li2017DailyDialogAM} dataset\footnote{Some sentences in the corpus are very long, we use the \textbf{nltk} toolkit to cut the long sentence.}. The format of the conversation and the details of the datasets are shown in Table \ref{tab:2}. The details of generated graph and datasets are shown in Table \ref{tab:1}.

\begin{table}[t]
    \begin{center}
        \resizebox{0.5\textwidth}{!}{
            \begin{tabular}{|c|l|}
            \toprule[2pt]
            \hline
            \multirow{5}{*}{Context} & A: What do you like to do with your free time?              \\ \cline{2-2} 
                                                  & B: Study English.                                           \\ \cline{2-2} 
                                                  & A: You mean you like to study english?                      \\ \cline{2-2} 
                                                  & A: That's weird. Why?                                       \\ \cline{2-2} 
                                                  & B: It gives me great satisfaction.                          \\ \hline
            \multirow{2}{*}{Reply}                & A: Wow, it wouldn't give me any satisfaction. \\ \cline{2-2} 
                                                  & A: It's a very hard work for me.                            \\ \hline \bottomrule[2pt]
            \end{tabular}
        }
        \caption{An example of the conversation in the modified dailydialog dataset. It should be noted that $A$ utters two responses at a time which is different from the existing datasets.}
        \label{tab:2}
    \end{center}
\end{table}

\subsection{Metrics}
We measure the performance of baselines and the proposed models from 2 aspects: (1) Language quality of responses; (2) Accuracy of the timing of talking.
As for the language quality of responses, we apply the human judgments and these automatic evaluations:
\begin{itemize}
    \item BLEU \cite{Papineni2001BleuAM}: We use the BLEU metric, commonly employed in evaluating the open-domain dialogue systems \cite{Li2015ADO, chen2017survey}.
    \item PPL: Perplexity is another commonly metric to evaluate the quality of the responses \cite{Zhang2019ReCoSaDT}, the lower the perplexity the better the quality.
    \item BERTScore \cite{Zhang2019BERTScoreET}: \cite{Liu2016HowNT} demonstrated that BLEU metric is not powerful for measuring the dialogue systems, 
    In order to make the results more reliable, we apply the state-of-the-art embedding-based metric BERTScore.
    \item Distinct-1(2): \cite{Li2015ADO} proposed the two metrics which measure the degree of diversity of the responses by calculating the number of distinct unigrams and bigrams.
\end{itemize}
For human judgments, given 100 randomly sampled context and their generated responses, three annotators (all CS majored students) give the comparsion between the proposed model and baselines, e.g. \textit{win}, \textit{loss} and \textit{tie} based on the coherence of the generated response with respect to the contexts. For exmaple, the \textit{win} label means the generated response of proposed model is better than baselines.

The decision of timing of talking is a binary classification task. Here we apply the \textit{Accuracy} and \textit{Macro}-F1 to measure the performance of the decision module.

\begin{table*}[]
    \small
    \begin{center}
        \resizebox{\textwidth}{!}{
            \begin{tabular}{|l|c|c|c|c|c|c|c|c|c|c|}
                \toprule[2pt]
                \hline
                \multicolumn{1}{|c|}{\textbf{Models}} & \textbf{PPL} & \textbf{BLEU1} & \textbf{BLEU2} & \textbf{BLEU3} & \textbf{BLEU4} & \textbf{BERTScore} & \textbf{Dist-1} & \textbf{Dist-2} & \textbf{Acc} & \textbf{Macro-F1} \\ \hline
                \textbf{HRED} & 63.06 & 0.0774 & 0.0486 & 0.0427 & 0.0399 & 0.8308 & 0.0427 & 0.1805 & - & -  \\ \hline \hline
                \textbf{HRED-CF} & \textbf{60.28} & \textbf{0.0877} & \textbf{0.0605} & \textbf{0.0547} & \textbf{0.0521} & 0.8343 & 0.0327 & 0.1445 & 0.6720 & 0.6713 \\ \hline
                \textbf{W2T-GCN} & 62.18 & 0.0844 & 0.0579 & 0.0521 & 0.0504 & 0.8342 & 0.0237 & 0.1028 & 0.6835 & 0.6741 \\ \hline
                \textbf{W2T-GAT} & 60.89 & 0.0746 & 0.0519 & 0.0477 & 0.0461 & 0.8320 & 0.0017 & 0.0029 & 0.7137 & 0.6686 \\ \hline \hline
                \textbf{W2T-GGAT} & 64.46 & 0.0816 & 0.0559 & 0.0504 & 0.0480 & \textbf{0.8344} & 0.0552 & 0.2651 & 0.7348 & 0.6936 \\ \hline
                \textbf{W2T-DGGNN} & 64.59 & 0.0796 & 0.0548 & 0.0496 & 0.0473 & 0.8328 & \textbf{0.0623} & \textbf{0.2790} & \textbf{0.7192} & \textbf{0.7015} \\ \hline
                \bottomrule[2pt]
            \end{tabular}
        }
        
        \resizebox{\textwidth}{!}{
            \begin{tabular}{|l|c|c|c|c|c|c|c|c|c|c|}
                \hline
                \multicolumn{1}{|c|}{\textbf{Models}} & \textbf{PPL} & \textbf{BLEU1} & \textbf{BLEU2} & \textbf{BLEU3} & \textbf{BLEU4} & \textbf{BERTScore} & \textbf{Dist-1} & \textbf{Dist-2} & \textbf{Acc} & \textbf{Macro-F1} \\ \hline
                \textbf{HRED} & 21.91 & 0.1922 & 0.1375 & 0.1266 & 0.1235 & \textbf{0.8735} & 0.0573 & 0.2236 & - & -   \\ \hline \hline
                \textbf{HRED-CF} & \textbf{20.01} & \textbf{0.1950} & 0.1397 & \textbf{0.1300} & \textbf{0.1270} & 0.8718 & 0.0508 & 0.2165 & 0.8321 & 0.8162 \\ \hline
                \textbf{W2T-GCN} & 23.21 & 0.1807 & 0.1262 & 0.1179 & 0.1160 & 0.8668 & 0.0437 & 0.1887 & 0.8273 & 0.8149 \\ \hline
                \textbf{W2T-GAT} & 21,98 & 0.1724 & 0.1181 & 0.1109 & 0.1129 & 0.8645 & 0.0077 & 0.0187 & 0.7290 & 0.7024 \\ \hline \hline
                \textbf{W2T-GGAT} & 21.12 & 0.1909 & 0.1374 & 0.1276 & 0.1249 & 0.8712 & \textbf{0.0698} & \textbf{0.3319} & 0.8281 & 0.8168 \\ \hline
                \textbf{W2T-DGGNN} & 21.52 & 0.1944 & \textbf{0.1399} & 0.1298 & 0.1269 & 0.8721 & 0.0673 & 0.3296 & \textbf{0.8369} & \textbf{0.8248} \\ \hline
                \bottomrule[2pt]
            \end{tabular}
        }
        \caption{Automatic evaluation results of all the models on Ubuntu and modified dailydialog test dataset (The table above is the results of ubuntu dataset, and the below is dailydialog). Baseline \textbf{HRED} cannot control the timing of talking and the \textit{accuracy} and \textit{macro}-F1 of it is empty. All the best results are shown in bold.}
        \label{tab:4}
    \end{center}
\end{table*}

\subsection{Baselines and Proposed models}
We choose the baselines as following:
\begin{itemize}
    \item HRED \cite{Li2016APN}: \cite{Li2016APN} propose the model for generating the dialogue responses with user embedddings. 
    We modified it by adopting the HRED architecture. Specifically, we concatenate the user embedding and the hidden state of the HRED context encoder \cite{Li2016APN} (Two speakers in the conversation). 
    It should be noted that the baseline is trained by the modified dataset mentioned above.
    \item HRED-CF: We add the same decision module in the proposed model for the HRED baseline and propose a powerful baseline called HRED-CF.
    \item W2T-GCN: To test the effectiveness of the proposed double-gated graph nerual networks, we replace the double-gated graph neural network with the GCN \cite{Kipf2016SemiSupervisedCW}. 
    \item W2T-GAT: Replace the double-gated graph neural network with GAT \cite{Velickovic2017GraphAN}.
    \item W2T-GGAT: Self-attention mechanism is another feasible method to detect the relative context. Here we replace the first gated mechanism with the self-attention mechaism (implemented by GAT) and construct a strong baseline W2T-GAT.
    \item W2T-DGGNN: W2T-DGGNN is our proposed double-gated graph neural network.
\end{itemize}

\subsection{Parameter Settings}

The parameter settings of our proposed models can be found in Table \ref{tab:4}. We implement the graph neural networks by using PyG \cite{fey2019fast}. For all the models, we adopt the early stopping to avoid the overfitting.

\begin{table}[!h]
    \small
    \begin{center}
        \begin{tabular}{|l|l|l|l|}
        \toprule[2pt]
        \hline
        \textbf{Param} & \textbf{Value} & \textbf{Param} & \textbf{Value} \\ \hline
        GNN layers $k$ & 3              & GRU $h$        & 500            \\ \hline
        Decision $h$ & 500/256/128/1  & learning rate  & 1e-3           \\ \hline
        Optimizer      & Adam           & weight decay   & 1e-4           \\ \hline 
        epochs         & 30             & dropout ratio  & 0.3            \\ \hline
        \bottomrule[2pt]
        \end{tabular}
        \caption{$h$ represents the hidden size. All the hidden size of the GRU are set as 500.}
        \label{tab:3}
    \end{center}
\end{table}

\begin{table*}[]
    \begin{center}
        \resizebox{\textwidth}{!}{
            \begin{tabular}{|c|c|c|c|}
            \toprule[2pt]
            \hline
            \textbf{Context}                                                                                                                                                                                                                                                                                                                                                                                                                           & \textbf{Ground-truth}                                                & \textbf{W2T-GGNN}                              & \textbf{HRED}                             \\ \hline
            \begin{tabular}[c]{@{}c@{}}A: I hear you bought a new house \\in the northern suburbs.\\ B: That's right, we bought it the same day \\we came on the market.\\ A: What kind of house is it?\\ B: It's a wonderful Spanish style.\\ A: Oh, I love the roof tiles on Spanish style houses.\\ B: And it's a bargaining.\\ B: A house like this in river side costs double the price.\\ A: Great, is it a two bedroom house?\end{tabular} & A: \textless{}slience\textgreater{}                         & A: \textless{}silence\textgreater{}   & A: Yes, it's a lot of money.     \\ \hline
            \begin{tabular}[c]{@{}c@{}}A: Oh, I'm starving.\\ A: I'd like to try some real Chinese cuisine.\\ A: What would you recommend, waiter?\\ B: Well, it depends.\\ B: You see, there are eight famous Chinese for instance,\\ the Sichuan cuisine, and the Hunan cuisine.\\ A: They are both spicy hot, I've heard.\\ B: That's right.\\ B: If you like hot dishes, you could try some.\end{tabular}                                   & A: They might be too hot for me.                            & A: Thanks for your advice.            & A: I have a good taste.          \\ \hline
            \hline
            \textbf{Context}                                                                                                                                                                                                                                                                                                                                                                                                                           & \textbf{Ground-truth}                                                & \textbf{W2T-GGNN}                              & \textbf{HRED-CF}                          \\ \hline
            \begin{tabular}[c]{@{}c@{}}A: Excuse me, I'd like to make a suit of this material.\\ B: Well, sir.\\ B: Can I take your measurements first?\\ A: Yes.\\ A: Should I take off my coat?\\ B: That will be helpful.\\ B: All right, thank you.\\ A: Please make a single-brested.\\ B: No, problem.\end{tabular}                                                                                                                     & \makecell[c]{B: If you will wait one moment, \\I'll make out your receipt.} & B: This's your receipt, here you are. & B: \textless{} silence\textgreater{} \\ \hline
            \bottomrule[2pt]
            \end{tabular}
        }
        \caption{The real examples in the dailydialog dataset.}
        \label{tab:6}
    \end{center}
\end{table*}

\subsection{Results}
\noindent \textbf{Generation Quality}: Table \ref{tab:4} shows the automatic evaluation results on the Ubuntu corpus and the modified dailydialog dataset. From Table \ref{tab:4}, we can make the following observations:
\begin{itemize}
    \item The strong baseline HRED-CF which explicity models the timing of talking is much better than HRED on almost all the metrics such as PPL and BLEU. Especially in terms of BLEU, HRED-CF exceeds HRED by up to 1.5\% on Embedding Average points. The experimental results demonstrate that controlling the timing of talking can significantly improve the quality the dialogue generation. The other two modified datasets shows the same conclusion which can be found in \textit{appendix}.
    \item Compared with the simple graph neural networks such as GCN and GAT (third and forth rows), our proposed gated neural network is much better. It indicates that the simply appling the graph neural networks is not enough and the gated mechanism is very important in modeling timing of talking.
    \item Compared with the strong baseline HRED-CF, the response quality of our proposed graph neural networks is weak. We attribute this phenomenon to the negative transfer problem \cite{Liu2019LossBalancedTW} during multi-task learning that the decision module influences the generation module. We leave this problem to the future research. But the quality of responses is still much better than HRED. It can be shown that although the BLEU scores, PPL and embedding average scores of GNN-based models are lower than HRED-CF, the diversity of the generation is better than strong baseline HRED-CF (distinct-1 and distinc-2).
    \item In order to test the effectiveness of the gated mechanism, we also replace the first gated mechanism with the self-attention attention mechanism and propose another strong baseline \textbf{W2T-GGAT} (fifth row). Except for PPL and \textit{accuracy}, \textbf{W2T-GGAT}'s performance is far worse than our proposed double-gated graph neural network. 
\end{itemize}

\begin{table}[h]
    \small
    \begin{center}
            \begin{tabular}{|c|c|c|}
                \toprule[2pt]
                \hline
                \textbf{Baseline} & \textbf{Human scores} & \textbf{Kappa} \\ \hline
                \textbf{HRED}     & 0.1111                      & 0.2983               \\ \hline
                \textbf{HRED-CF}  &                       &                \\ \hline
                \textbf{W2T-GCN}      &                       &                \\ \hline
                \textbf{W2T-GGAT}     &                       &                \\ \hline
                \textbf{W2T-GGNN}     &                       &                \\ \hline \bottomrule[2pt]
            \end{tabular}
        \caption{The human evaluation on the quality of the language on Dailydialog dataset.}
        \label{tab:7}
    \end{center}
\end{table}

As for the human evalution, the final results are 38.61\% / 12.87\% / 48.51\% for win, loss and tie which indicates that our proposed model can chat with human more naturally than HRED. 

\noindent \textbf{Decision Performance}: The performance of the decision module can be found in the last two columns in Table \ref{tab:4}. It can be shown that our proposed graph neural network models significantly outperforms the strong baseline HRED-CF by up to 5.77\% \textit{Accuracy} and 2.07\% \textit{macro}-F1. Furthermore, compared with the graph neural networks without gated mechanism, it can be found that out proposed model can achieve better performance on \textit{accuracy} and \textit{macro}-F1. It demonstrates that the gated mechanism is very important in modeling timing of talking. 

\subsection{Case Study}

As shown in Table \ref{tab:6}, 
to facilitate a better understanding of our proposed model, 
we provide some examples from dailydialog dataset. 
It can be found that proposed model can generate the more appropriate responses than HRED baseline.
Compared with the strong baseline HRED-CF and HRED, 
our proposed graph neural network can control the timing of talking accurately.

\section{Conclusion and Future work}
In this paper, we propose a new task for the generative multi-turn open-domain dialogue systems which explicitly controls the timing of talking. In order to model the timing of talking accurately for this task, we also propose a novel gated graph neural network. Extensive experiments demonstrate that: (1) Explicitly controlling the timing of talking during dialogue generation is benefit for the quality of the response; (2) Compared with the strong baselines, our proposed graph neural network model the timing of talking more accurately. Furthermore, we also release the source codes our models.

In future work, our research of this work are four-fold: (1) Leverage the reinforcement learning to improve the decision module; (2) Adopt other graph neural networks to model the timing of talking; (3) Alleviate the negative transfer issue of the multi-task learning; (4) Extend the dialogue generation with timing of talking to multi-party conversations.

\bibliography{anthology,acl2020}
\bibliographystyle{acl_natbib}


\end{document}